\documentclass[10pt,twocolumn,letterpaper]{article}

\usepackage{cvpr}
\usepackage{times}
\usepackage{epsfig}
\usepackage{graphicx}
\usepackage{amsmath}
\usepackage{amssymb}

\usepackage[table]{xcolor}
\usepackage{multirow}
\usepackage{booktabs}

\usepackage[pagebackref=true,breaklinks=true,letterpaper=true,colorlinks,bookmarks=false]{hyperref}

\cvprfinalcopy 

\def\cvprPaperID{7279} 

\ifcvprfinal\fi
\begin{document}

\title{RGB-based Semantic Segmentation Using Self-Supervised Depth Pre-Training}

\author{Jean Lahoud, Bernard Ghanem\\
King Abdullah University of Science and Technology (KAUST)\\
Thuwal, Saudi Arabia\\
{\tt\small \{jean.lahoud,bernard.ghanem\}@kaust.edu.sa}
}

\maketitle

\begin{abstract}

Although well-known large-scale datasets, such as ImageNet \cite{deng2009imagenet}, have driven image understanding forward, most of these datasets require extensive manual annotation and are thus not easily scalable. This limits the advancement of image understanding techniques. The impact of these large-scale datasets can be observed in almost every vision task and technique in the form of pre-training for initialization. In this work, we propose an easily scalable and self-supervised technique that can be used to pre-train any semantic RGB segmentation method. In particular, our pre-training approach makes use of automatically generated labels that can be obtained using depth sensors. These labels, denoted by HN-labels, represent different height and normal patches, which allow mining of local semantic information that is useful in the task of semantic RGB segmentation. We show how our proposed self-supervised pre-training with HN-labels can be used to replace ImageNet pre-training, while using 25x less images and without requiring any manual labeling. We pre-train a semantic segmentation network with our HN-labels, which resembles our final task more than pre-training on a less related task, \eg classification with ImageNet. We evaluate on two datasets (NYUv2 \cite{Silberman2012} and CamVid \cite{brostow2009semantic}), and we show how the similarity in tasks is advantageous not only in speeding up the pre-training process, but also in achieving better final semantic segmentation accuracy than ImageNet pre-training.
\end{abstract}

\section{Introduction}
One of the main goals in computer vision is to achieve a human-like understanding of images. This understanding has been recently represented in various forms, including image classification, object detection, semantic segmentation, among many others. All of these tasks are alleviated with large annotated datasets, especially after the emergence of deep learning techniques. The importance of such large-scale datasets \eg ImageNet \cite{deng2009imagenet}, can be easily observed in most deep learning methods, as nearly all techniques initialize from models pre-trained on these datasets.

\begin{figure}[t]
\begin{center}
   \includegraphics[width=\linewidth]{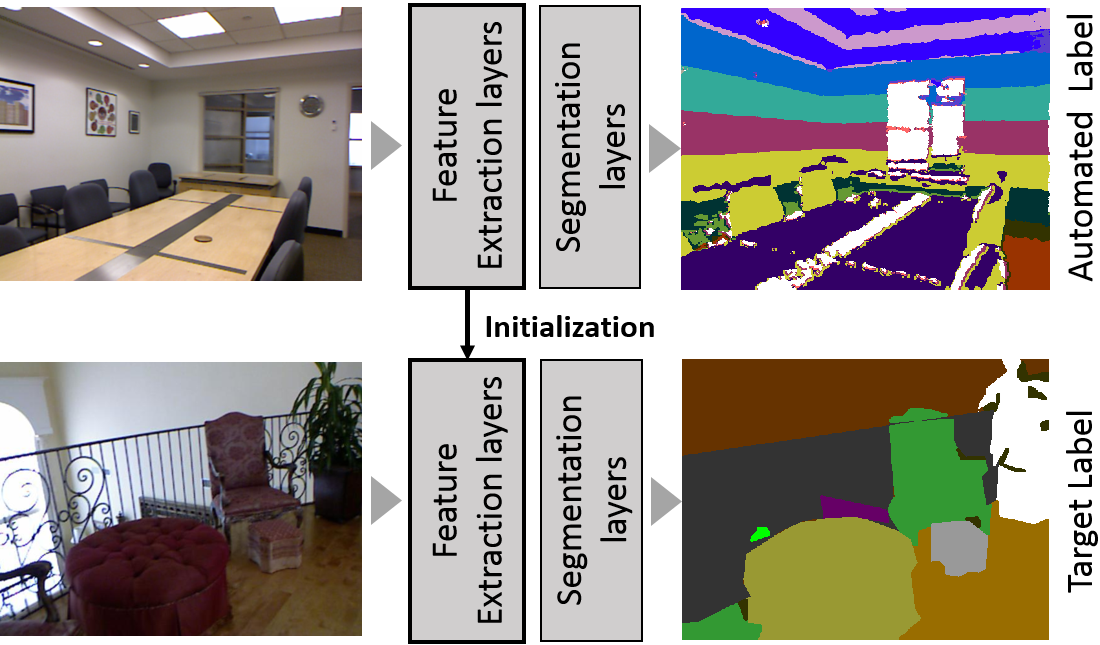}
\end{center}
   \caption{\small\textbf{Self-supervised learning with depth for pre-training.} Typically, a semantic segmentation model is initialized using another model pre-trained using hand labelled images (\eg ImageNet). We propose HN-labels that can be automatically generated from corresponding depth maps. These labels can be used for pre-training while maintaining the final semantic segmentation performance.}
\label{fig:output}
\end{figure}

\begin{table*}
\setlength\arrayrulewidth{1pt}
\begin{center}
\rowcolors{1}{white}{gray!15}
\begin{tabular}{l|cccc}
Method Requirement & Scalability & Straightforward Modeling & Data Available at test & Transferable to RGB only\\
\hline
Labeled RGB & -- & ++ & ++ & ++\\
Labeled RGBD & -- & + & - & - \\
Synthetic & o & ++ & ++ & o\\
Self-supervised & ++ & -- & ++ & ++\\
Model Based & - & o & - & o\\
Ours & + & ++ & ++ & ++\\
\hline
\end{tabular}
\end{center}
\caption{\textbf{A comparison between various training requirements for semantic segmentation.} We show the advantages (+) of every method requirement as well as the disadvantages (-).  (o) implies unclear benefit to the corresponding specification. Our proposed method is easily scalable, and operates directly on RGB images at test time without the need for additional information. Model based methods require a library of known shapes, such as 3D CAD models.}\label{tab:comparison}
\end{table*}

Since annotation is tedious and cannot be easily scaled, it currently represents a bottleneck to the advancement of deep learning methods. In an attempt to address this limitation, several approaches have explored the potential to minimize the amount of manual work required prior to training. One of the main trends today is the use of synthetic datasets to create photo-realistic environments that are inherently labeled. Although synthetic images have become to some extent similar to images taken in real environments, generating these realistic-looking scenes still requires  considerable manual effort in most cases. Another promising research direction is self-supervised learning, as well as reinforcement learning. Such learning does not require annotated data, and hence loosens the tie between learning methods and available datasets. Nevertheless, these  techniques require correct modeling and might become computationally demanding.

This work aims at generating labels that \emph{(i)} can be automatically obtained without manual annotation, and \emph{(ii)} improve the task of semantic segmentation. Specifically, we use the depth channel supplied by RGBD sensors to extract patches that are labeled according to height and normal information only. We denote our labels by HN-labels, which we use to pre-train a semantic segmentation network in a self-supervised fashion. In this way, both our pre-training and training processes are closely related as both aim at semantically segmenting a given scene. Our method makes use of this depth information for pre-training only, as depth information need not be present at inference time.

We show how our HN-labels can easily replace large-scale annotations required for pre-training. An added advantage of our proposed technique is that it helps introduce new segmentation models not based on popular architectures such as VGG-16 \cite{Simonyan14c}, AlexNet \cite{NIPS2012_4824}, GoogleNet \cite{googlenet}, and ResNet \cite{he2016deep}. Our goal is to substantially reduce the amount of manual work required prior to training. Therefore, we formulate the pre-training as a self-supervised learning process. The only manual work needed is the collection of aligned RGBD images, which is straightforward,  not time consuming, and does not require manual semantic labeling. Figure \ref{fig:output} shows the end target of our method: to replace hand-labeled datasets used in initializing semantic segmentation networks with a dataset that can be easily collected and automatically labeled, without significant impact to segmentation performance. 


We formulate the initialization as an HN-labels segmentation task, which is closely related to semantic segmentation. 
The HN-labels represent different classes that are automatically formed based on grouping normal angles and height relative to the floor plane. Learning to segment the HN-labels classes enables mining of local semantic information. We show how initializing from a closely related task is much more efficient than initializing from less related tasks. Predicting heights and normals from a single RGB image is a by-product of our method and is not the main goal. Similar to depth estimation, height can take continuous values in 3D which makes the estimation task harder. Nevertheless, indoor scenes usually have structured environments in which common objects generally have similar heights. Unlike depth, it is important to note that height are independent of the camera viewpoint. 


Table \ref{tab:comparison} shows a comparison between the various training requirements of semantic segmentation techniques. Our approach aims at providing labels that can be easily scaled, do not require additional complex modeling, and can be easily used to label any given RGB image.

\section{Related Work}
\begin{figure*}[ht!]
\begin{center}
\includegraphics[width=\linewidth]{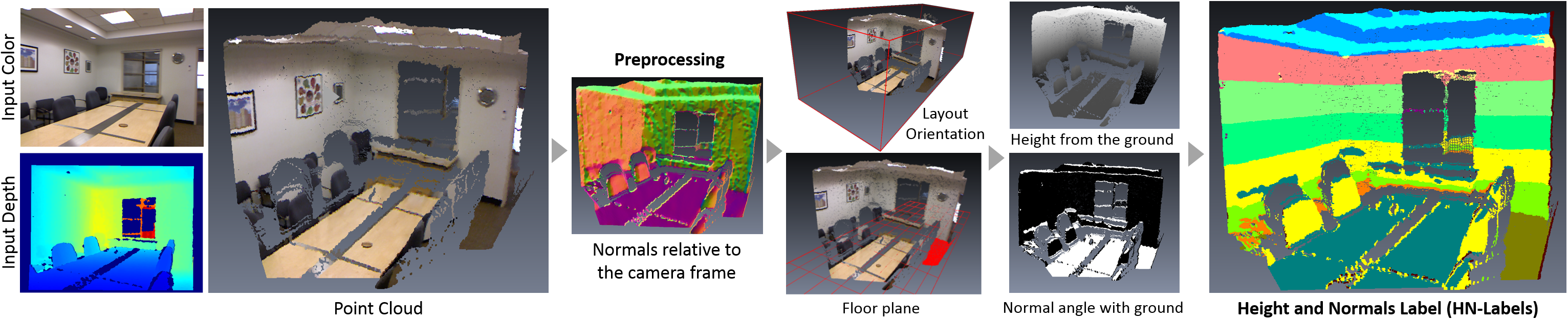}
\end{center}
\vspace{-10pt}
   \caption{\textbf{General overview of our method.} Given an RGB image along with its corresponding depth image, we compute the normals on the generated point cloud. We then estimate the room orientation, and extract the floor plane. Height and normal angles relative to the floor are then used to generate our HN-labels.}
\label{fig:overview}
\end{figure*}

Since this work aims at providing better semantic segmentation based on transferring knowledge from a closely related and easily scalable self-supervised task, we provide a literature review on three related topics: \textbf{(i)} semantic segmentation, \textbf{(ii)} multi-task and transfer learning, and \textbf{(iii)} self-supervised and scalable learning.

\vspace{8pt}
\noindent\textbf{Semantic Segmentation:} The recent success of deep learning techniques in image classification and object detection have motivated researchers to explore the task of semantic segmentation with deep learning. This could only be achieved with pixel-level labeling provided by large-scale datasets, such as PASCAL VOC \cite{everingham2010}, MS COCO \cite{lin2014}, Cityscapes \cite{cordts2016}, CamVid \cite{brostow2009}, and KITTI \cite{geiger2013}. Most recent methods benefit from existing classification CNNs (Convolution Neural Networks) by adapting them into ones that can output pixel labeling instead of classification scores. An early method is the Fully Convolution Network (FCN) network \cite{long2015fully}, which replaced the fully connected layers of popular classification networks with what was later referred to as `deconvolutions' to produce pixel labeling. A similar approach was proposed with the SegNet architecture \cite{badrinarayanan2015segnet}, which also utilized convolution layers from a classification network (encoder), but introduced a different upsampling mapping (decoder). The upsampling was done through multi-stage learnable `deconvolutions' with upsampling indexed similar to max-pooling indices of the encoder part. These techniques, along with numerous others such as \cite{chen2016deeplab,chen2017rethinking,zhao2017pyramid,yu2015multi}, require large-scale datasets in training and can only be improved with more groundtruth labeling. 

Indoor scenes benefit from the use of off-the-shelf RGBD sensors that measure coupled depth and color information. Multiple datasets were collected using such sensors and were labeled for pixel-level semantic segmentation, including NYU-D v2 \cite{Silberman2012} and SUNRGBD \cite{song2015sun}. Encoding the depth cue as an input to neural networks is not straightforward, and multiple approaches have been proposed. Gupta \etal \cite{gupta2014learning} proposed to encode depth as additional channels in the form of horizontal disparity, height above ground, and the angle between the local surface normal and the gravity direction (HHA). This leads to better segmentation compared to using the color channel alone. However, this method, similar to all RGBD-based techniques, requires the presence of depth information at inference time.

\vspace{8pt}
\noindent\textbf{Transfer Learning and Multi-task Learning:} In practice, most CNN approaches do not train their models from scratch. Instead, learned feature extractors are transferred  \cite{sharif2014cnn} from models trained on large-scale datasets, such as ImageNet \cite{deng2009imagenet}. Other forms of transfer learning can be done by pre-training on synthetic datasets \cite{ros2016synthia,gaidon2016virtual}. Transfer learning can be viewed as a special case of multi-task learning, where tasks are sequentially learned instead of being jointly learned. Numerous works train a network to perform multiple tasks simultaneously. The approach proposed by \cite{eigen2015predicting} jointly predicts depth, surface normals, and semantic labels with a multi-scale network. Although these task are closely related, labels for all tasks need to be available for training. The work presented in \cite{misra2016cross} introduces a principled way to share activations from multiple networks. The sharing unit, known as cross-stitch, can be trained end-to-end to choose between sharing the layers or separating them at task-specific activations. Other work \cite{kendall2017multi} proposes a systematic way to combine multiple loss functions of multiple tasks by considering the uncertainty of each task.

\vspace{8pt}
\noindent\textbf{Self-Supervised Learning:} The technique presented in \cite{zeng2017multi} proposes a self-supervised method to generate a large labeled dataset without the need for manual labeling. Labels of known objects were generated by manipulating the camera around a controlled environment, in which object locations are known. The generated dataset was then used to train a robot to better perform in picking tasks. A main drawback of this technique is that the dataset was generated in a controlled environment with known camera locations and also required human involvement in arranging objects. Other self-supervised techniques aim at regressing depth from videos \cite{godard2016,zhou2017,vijayanarasimhan2017}. Hence, these techniques are good at estimating depth especially in outdoor scenes in the context of autonomous driving. These applications mainly focus on objects located on the road. This greatly differs from how object are arranged in indoor scenes. Since we model our pre-training as a self-supervised process, the amount of RGBD data that is available for pre-training is large and can be easily scaled up. One can easily append training data with personal data collected with RGBD sensors in any environment of interest. Self-supervision for representation learning was also explored in \cite{noroozi2016unsupervised, larsson2017colorization}, in which proxy tasks were proposed using the color images. Although our proposed proxy segmentation task requires depth data, this additional data provides information that cannot be learned from the image alone. Techniques, which only require color, are a good alternative, but since they are not provided with inaccessible information, their performance is shy of ImageNet pre-training for the same task.

\vspace{8pt}
\noindent\textbf{Contributions.} We propose a self-supervised technique for the task of height and normal estimation from a single RGB image. The network trained on this task can be used to initialize a semantic segmentation network. We show how this self-supervised pre-training strategy can be easily scaled and is more efficient than initializing from networks trained on different fully supervised tasks. Our experiments show that semantic networks initialized using our proposed self-supervised method outperform ones initialized from networks trained on large-scale datasets annotated for a different task. 

\section{Methodology}
Our method aims at overcoming the limitations of semantic segmentation techniques when training on a small dataset or when training from scratch. An overview of our method is depicted in Figure  \ref{fig:overview}. In the following sections, we detail our process in choosing the labels, generating the dataset, and using these labels to pre-train a given network.

\subsection{Alternative Category Representation}\label{sec:label_choice}
The motivation to this work is to look at the depth channel as a source for labeling RGB images. RGBD sensors are used to easily collect aligned color and depth image pairs. Our aim is to automatically generate labels that are not necessarily designed for the target task (semantic segmentation), but could afford helpful information for it. Successfully transforming depth into labels makes this method capable of generating \emph{free} labels as compared to the tedious, time consuming, RGB pixel labeling needed for semantic segmentation. The first intuitive use of depth is to train a network to directly regress the depth from a single RGB image. However, depth requires complex knowledge of the environment and varies with camera location/viewpoint.

With the complexity of indoor scenes, regressing depth necessitates a high capacity network and can be tedious to train. On the other hand, with additional effort, one can transform depth into height above ground, which is view-independent. Taking the ground plane as reference, one can also transform normals relative to the camera viewpoint, which depend on camera angle, to view-independent normal angles relative to the ground plane.

\begin{figure}[!h]
\begin{center}
\includegraphics[width=\linewidth]{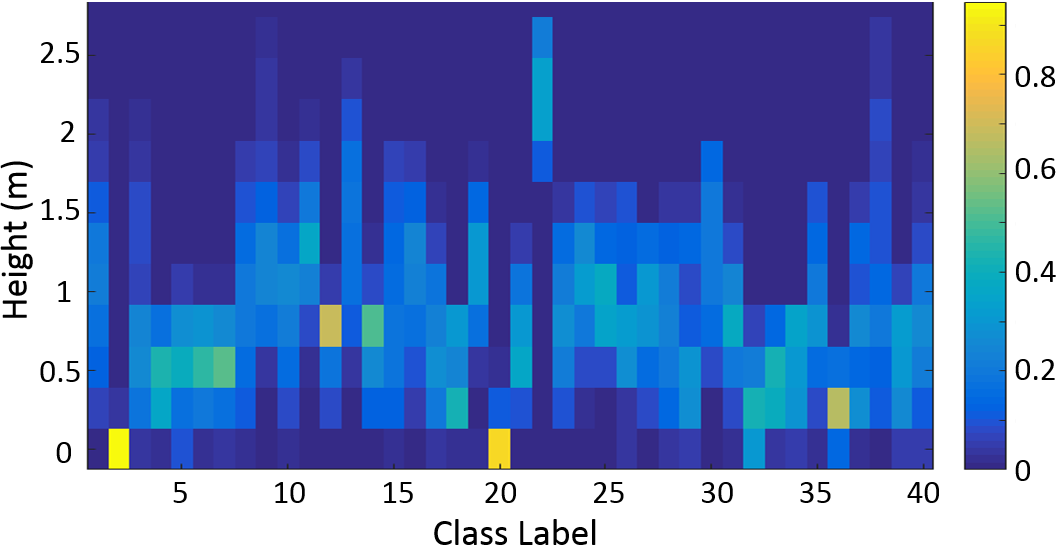}
\end{center}
\vspace{-5pt}
   \caption{\textbf{Object height distribution.} This colormap shows the distribution of height for various object classes in NYUv2 \cite{Silberman2012}. Colors represent the fraction of 3D points of the same label that are located at a particular height bin.} 
\label{fig:height_stats}
\end{figure}

Our method uses height and normal information as the main precursor for pre-training. We choose to bin the height into $n_h$ bins (or levels) and normals into $n_n$ bins and train a model to output the correct normal and height level per pixel. Since it learns to generate features specific to segmentation, this pre-training is closer to our end goal of semantic segmentation than to pre-train with models trained on other tasks, \eg image classification. In fact, one can think of height bins as semantic segmentation with object part specific labels: chair and table lower legs, lower part of wall and door, table top, and sofa back, \textit{etc}. This type of grouping consistently clusters objects and their parts into specific bins. To highlight this point, Figure \ref{fig:height_stats} shows the height distribution of 40 indoor objects on a subset of NYUv2 dataset \cite{Silberman2012}. For every object, we show the percentage of its occurrence in every height bin. As expected, objects like floor (label 2) and floor mat (label 20) only occur within the lowest height bin, whereas ceiling (label 22) occurs within the highest bins. Other objects almost always occur in the same bin, such as bathtub (label 36) and counter (label 12). Furthermore, when considering other object classes, there exists an overall trend in height, with object parts falling in specific height bins. In light of these observations, training a network to distinguish between different height levels will inherently train it to distinguish between some object classes and/or their parts. This information is expected to be useful for the task of semantic segmentation.


\subsection{Automated Label Collection}\label{sec:data_collec}
\noindent\textbf{Floor Identification:} Our method mainly relies on the floor plane as a reference for the proposed labels. To identify a possible floor plane, we first use the  camera intrinsic parameters to map the depth image into a 3D point cloud. Next, we compute the normals at each 3D point using the method of \cite{rosman2014augmented}. We then use the normals to estimate the scene layout orientation using the method of \cite{ghanem2015robust}. Once the orientation is obtained, the room is rotated to make the ground normal aligned with the upward direction. Afterwards, we choose the ground location as the first percentile of points along the upward direction. The first percentile is chosen instead of the minimum in order to make the method more robust to sensor errors especially at far distances. Our method takes into account that in most indoor images captured by RGBD sensors, a part of the ground is visible. In order to account for cases where the ground is not visible, we discard scenes in which the majority of the normals from points on the hypothesized floor plane (first percentile) are not aligned with the upward direction.

\begin{figure}[!h]
\begin{center}
\includegraphics[width=\linewidth]{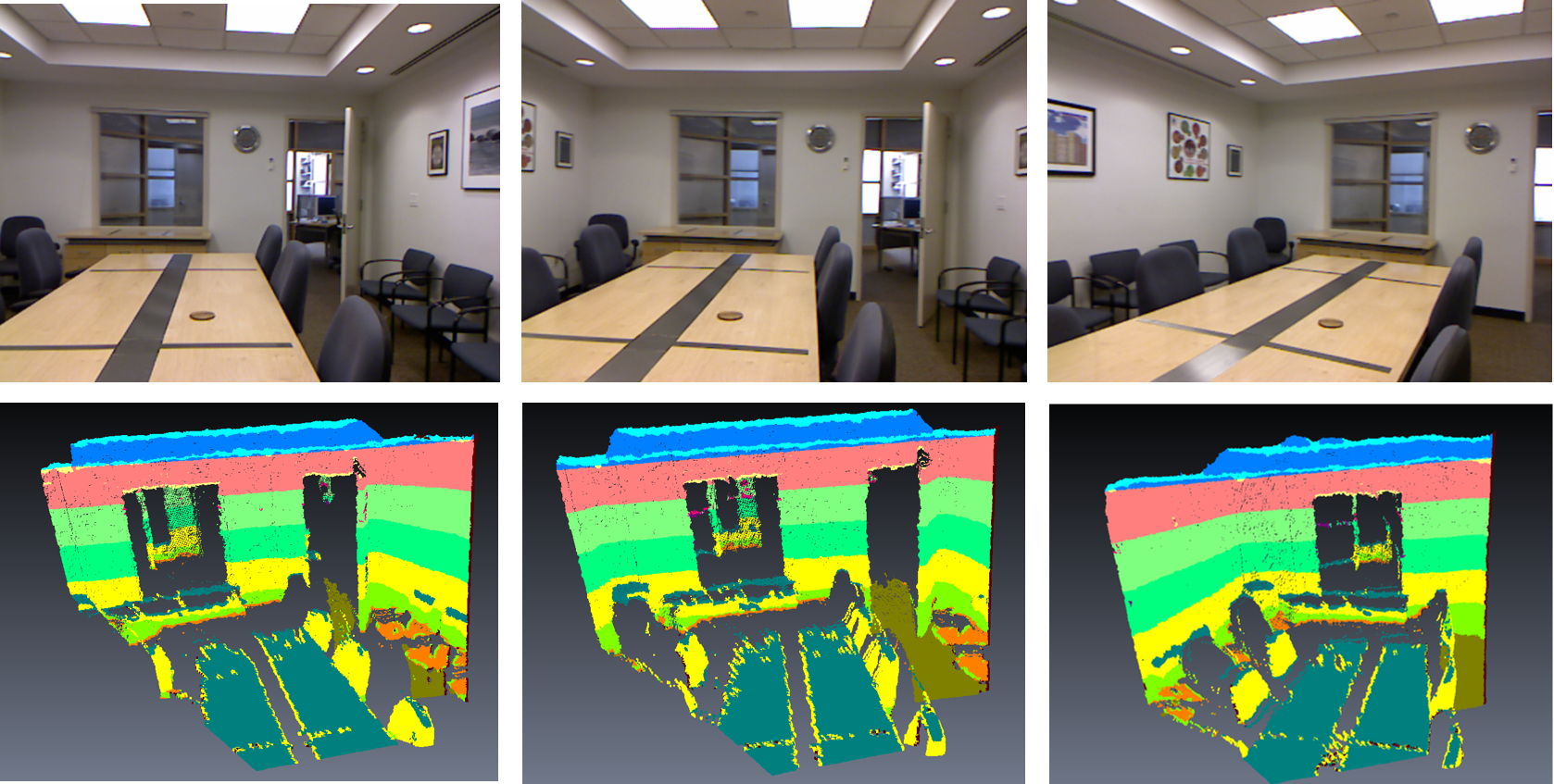}
\end{center}
   \caption{\textbf{HN-labels for different images in the same room.} Our proposed labels are view-independent and are consistent among different viewpoints.}
\label{fig:view_indep}
\end{figure}

\noindent\textbf{HN-Label Generation:} To transform depth into height, we calculate the distance between each 3D point and the floor plane. As for the normals, we compute the angle between every 3D point normal and the floor's upward direction. We only use one angle to keep labels consistent over multiple viewpoints. Normals and heights are then binned, and we define our proposed alternative categories as all possible combinations of the normal and height bins. Figure \ref{fig:view_indep} shows our proposed labels for the same scene from different viewpoints. Since we choose our labeling scheme to be view-independent, objects maintain the same labels throughout the scene recording. 

\subsection{Training with HN-Labels}\label{sec:train_alter}
Labels collected from height and normal information can be used to train the same network architecture used for semantic segmentation. Specifically, we choose the SegNet architecture \cite{badrinarayanan2015segnet} for this purpose, with an encoder network identical to the 13 convolutional layers in the VGG16 network \cite{Simonyan14c}, as well as, the DeepLabv3 architecture \cite{chen2017rethinking} with ResNetv2 layers \cite{he2016identity}. Note that our method can be easily implemented with any other architecture and that the  number of labels for semantic segmentation need not be similar to that of pre-training with HN-labels. In what follows, we will reference this trained network as the HN-network. 


Once we train our HN-network with labels that are self-supervised, we need to transfer the representation to another network that learns to segment RGB images. In both the SegNet and DeepLabv3 architectures, the only difference between the two networks is the number of object classes, \ie  the number of outputs from the last convolution layer. Transfer learning can be done in two ways. The first way is to initialize all the layers of the semantic segmentation network with what was learned in the first network, and then finetune all the layers based on the manually labeled segmentation dataset. Another way of transfer learning is to keep some of the first layers unchanged, especially when the dataset used for pre-training is much larger than the dataset used for finetuning. This is motivated by the idea that  generic information is usually shared at early layers.

\begin{figure}[t]
\begin{center}
\includegraphics[width=\linewidth]{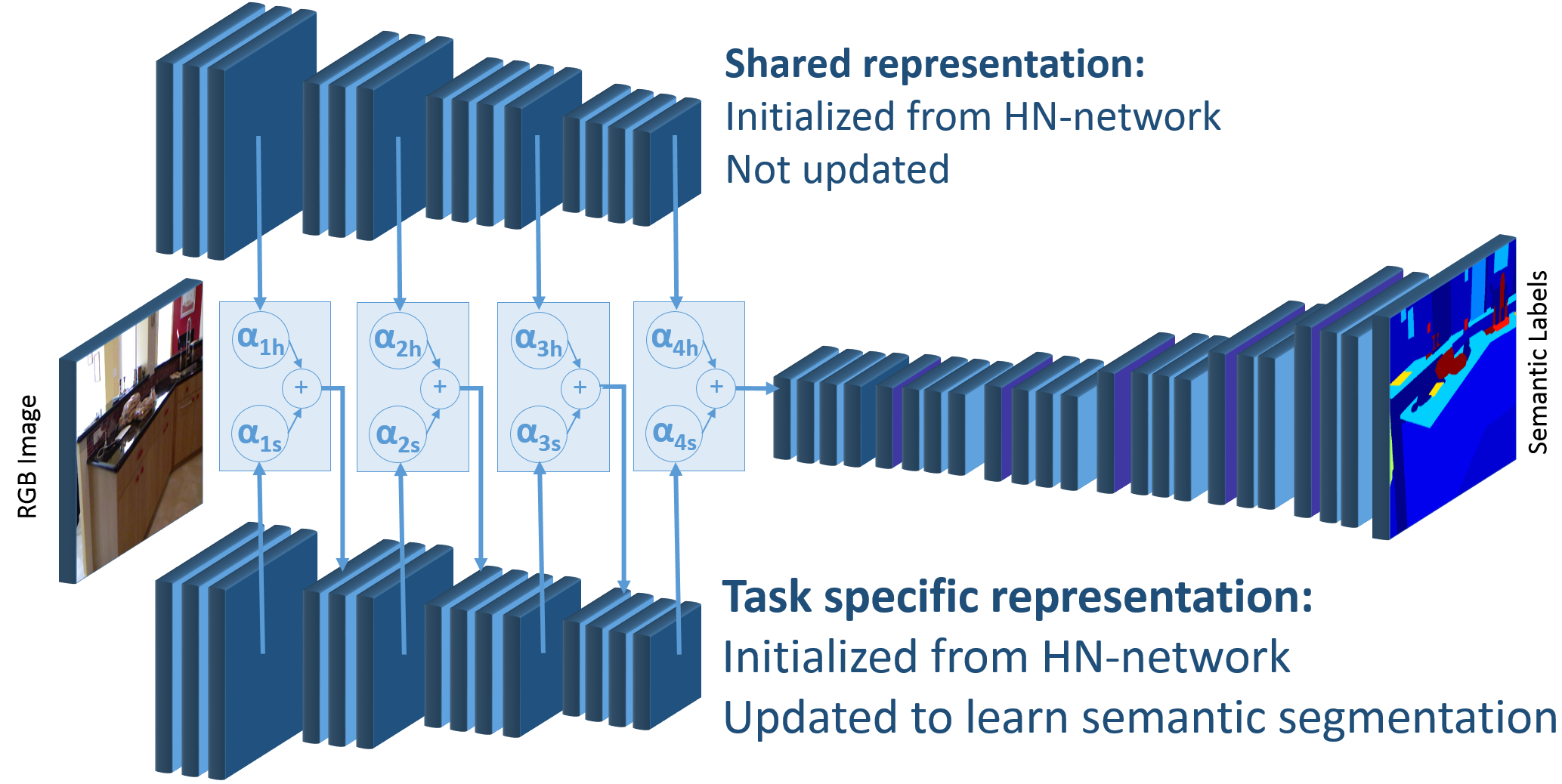}
\end{center}
   \caption{\textbf{Proposed fusion of HN-network and semantic segmentation network (Cross-Stitch).} The HN-network is considered as a shared representation whereas the semantic segmentation network is task specific). At every max-pooling layer, the network gets to choose between the pre-trained layers and the task specific layers.}
\label{fig:cross}
\end{figure}

Choosing to finetune only part of the network raises an important question: how many layers should be fixed and how many should be finetuned? Instead of manually selecting which layers to share and which layers to be task-specific, one can model the problem as a multi-task learning problem, in which sharing can be learned between the networks. This is mainly inspired by the work in \cite{misra2016cross}. Our joint network architecture, named cross-stitch \cite{misra2016cross}, is shown in Figure \ref{fig:cross}. The shared representation part of the network is extracted from a network that was trained to output HN-labels, whereas the task-specific part learns to semantically segment. The shared representation part is kept fixed, reasoned by the large data that can be used to train that part. Sharing of layers is allowed at the first 4 max-pooling layers. We denote the output at every max-pooling by $L_{xs}$ and $L_{xh}$, where $x$ represents the index of the max-pooling layer. $L_{xs}$ refers to the output generated in the ``semantic segmentation network", whereas $L_{xh}$ refers to output from the HN-network. At every max-pooling layer, both representations are merged into $$L = \alpha_{xh}L_{xh} + \alpha_{xs}L_{xs}$$ and $L$ is used as an input to the subsequent layer of the semantic segmentation part of the network.

The final target is to achieve a higher accuracy in semantic segmentation only, especially when semantic labels are scarce. This differs from \cite{misra2016cross}, which aims at achieving a better joint accuracy in multiple tasks.

\noindent \textbf{Implementation Details: } Our method is implemented using the Caffe framework \cite{jia2014caffe} for the SegNet architecture and Tensorflow \cite{abadi2016tensorflow} for the DeepLabv3 architecture, both running on Titan X and Titan Xp GPUs. We use the same image input and output size for all of our networks. In the cases where depth sensors output a different size image, we compute the height on the original size image, then crop and reshape the color image with bilinear interpolation, while using nearest-neighbour interpolation for the labels. As for the training loss, we use the cross-entropy loss \cite{long2015fully} in all our networks. 


\section{Experiments}

\begin{figure}
\begin{center}
\includegraphics[width=\linewidth]{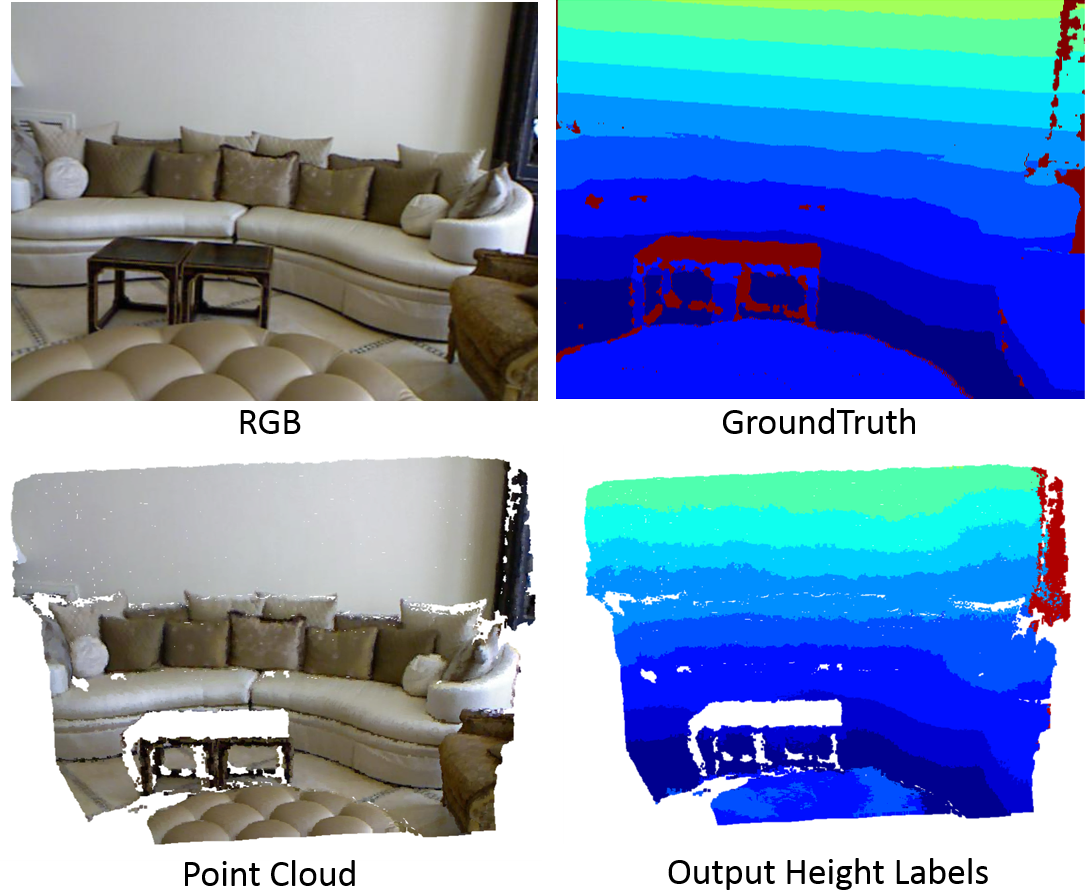}
\end{center}
   \vspace*{-3mm} 
   \caption{\textbf{Example of the output height labels obtained from the HN-network.} The labels are inpainted on the corresponding 3D point cloud. Groundtruth height labels are also shown.}
\label{fig:height_eg}
\end{figure}

\subsection{Datasets}
We generate our alternative label dataset from the unlabeled set of NYUv2 \cite{Silberman2012} dataset, which comprises more than 400K frames of indoor RGBD video sequences.

To evaluate the effect of our pre-training on semantic segmentation, we use the 1449 segmented frames of NYUv2 that are split into train and test, with the class label mapping coming from \cite{gupta2013perceptual}. We also perform experiments on the Cambridge-driving Labeled Video Database (CamVid) \cite{brostow2009semantic}, which contains scenes from the perspective of a driving automobile.

We use the following three conventional evaluation metrics: \textbf{(i)} global accuracy, \textbf{(ii)} class average accuracy, and \textbf{(iii)} mean intersection over union (IoU). Global accuracy is computed as the ratio between the number of correctly labeled pixels and the total number of pixels. The class average accuracy represents a scoring measure that weighs all classes similarly. The mean IoU also presents a measure that equally accounts for all classes and does not favor high accuracies in frequently occurring object classes. 
\begin{table}
\begin{center}
\setlength\tabcolsep{4pt}
\begin{tabular}{l|c|c|c}
\toprule
Pre-training Method & Global acc & Avg acc & mIoU\\
\midrule
no pre-training & 38.74 & 13.73 & 8.78 \\
height only & 50.66 & 24.27 & 16.35 \\
height 5 bins + normals & 51.98 & 25.28 & 17.38 \\
height 10 bins + normals & \bf{52.92} & \bf{26.16} & \bf{18.24} \\
\bottomrule
\end{tabular}
\end{center}
\caption{\textbf{Ablation study.} This table shows the semantic segmentation network accuracy on NYUv2 when pre-trained from different versions of our proposed HN-network.}\label{tab:ablation}
\end{table}

\begin{figure}
\begin{center}
\includegraphics[width=\linewidth]{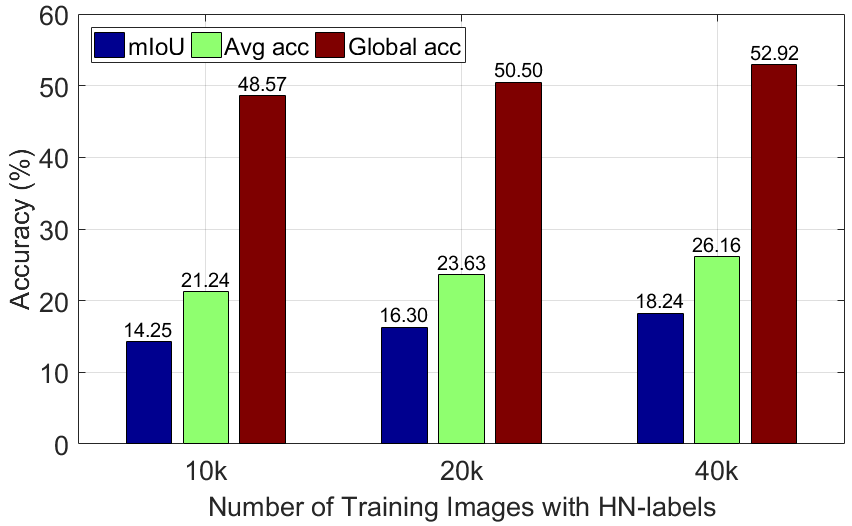}
\end{center}
   \vspace*{-3mm} 
   \caption{\textbf{Effect of dataset size.} Semantic segmentation network accuracy on NYUv2 increases when pre-trained from a larger number of training images with HN-labels.}
\label{fig:dataset_size}
\end{figure}

\begin{table*}
\begin{center}
\setlength\tabcolsep{3.5pt}
\begin{tabular}{c|l|ccccccc|c|c|c}
\toprule
Arch & Pre-training Method & Wall & Floor & Cabinet & Bed & Bookshelf & Sofa & Dresser & Avg acc & mIoU & Global acc\\
\midrule
\parbox[t]{2mm}{\multirow{4}{*}{\rotatebox[origin=c]{90}{SegNet}}}&No pre-training & 77.57 & 74.91 & 42.36 & 13.25 & 6.58 & 22.42 & 3.17 & 13.73 & 8.78 & 38.74\\
&CIFAR100 pre-train  & 66.33 & 66.96 & 19.84 & 0.03 & 0.01 & 0.05 & 0 & 10.70 & 6.23 & 31.90 \\
&ImageNet pre-train & 81.54 & 86.1 & \bf{58.86} & 37.76 & 35.05 & \bf{38.25} & 4.35 & 25.40 & 17.34 & 50.58\\
&HN pre-training & \bf{83.26} & \bf{90.63} & 58.81 & \bf{58.40} & \bf{38.79} & 37.05 & \bf{14.24} & \bf{26.16} & \bf{18.24} & \bf{52.92} \\
\midrule
\parbox[t]{2mm}{\multirow{3}{*}{\rotatebox[origin=c]{90}{\small{DeepLab}}}}&No pre-training & 18.70 & 66.72 & 60.78 & 22.30 & 20.70 & 15.24 & 12.49 & 24.77 & 6.73 & 31.41\\
&ImageNet pre-train & 28.85 & 74.69 & \bf{87.98} & \bf{64.40} & \bf{56.22} & 59.10 & 50.78 & \bf{57.84} & \bf{34.27} & 61.70\\
&HN pre-training & \bf{32.00} & \bf{77.64} & 87.45 & 60.15 & 55.67 & \bf{60.51} & \bf{53.69} & 56.04 & 33.49 & \bf{62.98} \\
\bottomrule
\end{tabular}
\end{center}
\caption{\textbf{Evaluation of our proposed method.} The accuracy on some of the most occuring object labels in NUYv2 is shown, as well as the global and average accuracies on all 40 classes. Our method is compared to three baselines. The first one does not use additional labels to pre-train, whereas the second and third baselines use manually annotated labels for pre-training.}\label{tab:results}
\end{table*}

\subsection{Training the HN-Network}\label{subsec:trainign HN}
To train our proposed HN-network using the generated alternative labels, we use 41K frames from NYUv2 sampled uniformly from all scenes. We use a fixed $n_n=2$ for normal bins, as indoor objects have surfaces that are mostly oriented either parallel or perpendicular to the floor plane. Our final label set contains $n_n \times n_h$ labels, with $n_h=10$, and we discard points that do not have depth information. Also, we use the softmax cross-entropy loss during training that discards pixels with unknown depth label. For all experiments, the input image size is $424$$\times$$560$, which enables us to only fit a batch of 4 images when training our HN-network. After 200K iterations in about 2 days, the training loss decreases to 4-5 times its initial value. We test our HN-network on NYUv2 test set. An example of the predicted height of the HN-network is shown in Figure \ref{fig:height_eg} and compared to the groundtruth. Clearly, the network is able to learn and predict the correct height bin, which is an essential prior to using the same representation for semantic segmentation.

\subsection{Training for Semantic Segmentation}
We use the trained HN-network to initialize our second network for semantic segmentation on RGB images, as described in Section \ref{sec:train_alter}.
\\

\begin{table}
\begin{center}
\begin{tabular}{l|c|c|c|c}
\toprule
Method / Epoch & 5 & 10 & 50 & 100\\
\midrule
No pre-training & 10.69 & 31.95 & 54.85 & 71.07 \\
CIFAR100 pre-train & 3.54 & 9.98 & 2.58 & 15.65\\
ImageNet pre-train & 24.97 & 27.1 & 42.4 & 85.87\\
HN pre-training & \bf{78.98} & \bf{81.70} & \bf{84.5} & \bf{86.03}\\
\bottomrule
\end{tabular}
\end{center}
\caption{\textbf{Evaluation on the CamVid dataset using SegNet architecture.} The global accuracy is shown at different number of training epochs} \label{tab:camvid}
\end{table}

\noindent \textbf{Ablation Studies: }
A comparison between different versions of our semantic segmentation method is presented in Table \ref{tab:ablation}. We fix all the learning hyperparameters and run our experiment with SegNet architecture for 40K iterations (equivalent to about 50 epochs on the NYUv2 dataset). We compare our results against training from scratch with weight filler from \cite{he2015delving}. Note that all measures are computed on 40 classes of NYUv2, which only contains 795 training images. Our network pre-trained with height labels achieved good performance, whereas adding the normal information boosted up the accuracy. 

We also study the effect of training size on the final semantic segmentation accuracy in Figure \ref{fig:dataset_size}. In this case, we fix the number of iterations for the HN-network and semantic segmentation training. Our proposed method achieves good accuracy even with relatively small number of images with HN-labels. Higher accuracy is achieved with more training images, thus, exploiting the diversity among images in the larger set.
\\

\noindent \textbf{Semantic Segmentation Evaluation: }
Here, we compare our results to training from scratch, as well as, pre-training from the ImageNet and CIFAR100 \cite{krizhevsky2009learning} classification datasets. Semantic segmentation results on NYUv2 dataset are shown in Table \ref{tab:results} for both the SegNet architecture \cite{badrinarayanan2015segnet} and the DeepLabv3 architecture \cite{chen2017rethinking}. All fine-tuning experiments were done for for 40K iterations ($\sim$ 50 epochs). If we compare the size of ImageNet to that of our alternative label dataset, ImageNet is about 25 times larger and requires more time to train. Nevertheless, initializing from our proposed dataset outperforms the ImageNet initialization for both architecture types. This shows that ability of our method to generalize to other architectures and backbones without requiring any expensive manual labelling. CIFAR100 initialization shows a degradation in accuracy over training from scratch (MSRA initialization \cite{he2015delving}). This degradation is not unusual in the deep learning field, where training with atypical data would produce results closer to undesirable local minima. 

Table \ref{tab:camvid} shows our experiments on the CamVid dataset using the SegNet architecture
with VGG16 encoder, which contains images collected from outdoor settings. Initializing from the HN-labels dataset is still beneficial in such a setting, mainly due to the similarity in the task. The main intuition is that training on HN labels has learned to describe low-level feature information for the visual content in the image. Nonetheless, there exist differences in the high-level information between indoor and outdoor scenes, and this is noticeable in the relatively lower improvement for the outdoor scenes. Also, our method achieves the final accuracy much faster than other methods. This shows that although most objects in the outdoor setting were not previously seen by our pre-training network, it has still learned to discriminate between different patches.  
\begin{table}[!ht]
\begin{center}
\begin{tabular}{l|c|c}
\toprule
Fixed Layers & ImageNet & Ours \\
\midrule
conv1 & 50.1 / 25.1  /16.9 & 54.1 / 27.9 / \bf{19.7}  \\
conv1-conv2 & 51.5 / 25.3 / 17.5 & 54.0 / 27.8 / 19.6  \\
conv1-conv3 & 51.5 / 25.3 / 17.4 & 52.9 / 26.9 / 18.7  \\
conv1-conv4 & 50.8 / 25.4 / 17.5 & \bf{54.4} / \bf{28.1} / {\normalfont 19.4}  \\
\bottomrule
\end{tabular}
\end{center}
\caption{\textbf{Effect of fixing layers relative to finetuning all layers.}(Global acc/Avg acc/mIoU). Although ImageNet pre-training leads to better performance when some layers are fixed, our proposed method would always outperform it in the same setting.}\label{tab:keep_layers}
\end{table}

We also try to initialize the segmentation network from our HN-network while keeping some layers intact. We finetune all other layers and stop at 40K iterations. The result on NYUv2 is shown in Table \ref{tab:keep_layers}, which shows how global accuracy varies with fixing more layers. As can be seen, fixing up to four layers improves the final accuracy. This result can be reasoned because of the significant difference between the dataset used for pre-training and the one used for finetuning. Nevertheless, this observation cannot be used as a general conclusion as best accuracies can be achieved by fixing different number of layers in different scenarios. This motivates our next experiment, which learns what to share among layers between the two networks.

\begin{table}
\begin{center}
\begin{tabular}{l|c|c|c}
\toprule
Method & Global acc & Avg acc & mIoU\\
\midrule 
ImageNet pre-training & 50.58 & 25.40 & 17.34 \\
ImageNet cross-stitch & 50.94 & 25.33 & 17.35 \\
Ours - No stitching & 52.92 & 26.16 & 18.24 \\
Ours cross-stitch & \bf{54.25} & \bf{27.83} & \bf{19.46} \\
\bottomrule
\end{tabular}
\end{center}
\caption{\textbf{Evaluation of the cross-stitching result.} The cross-stitch result is compared against no stitching and ImageNet pre-training.}\label{tab:results_cross}
\vspace{-10pt}
\end{table}

\begin{figure*}
\begin{center}
\includegraphics[width=\linewidth]{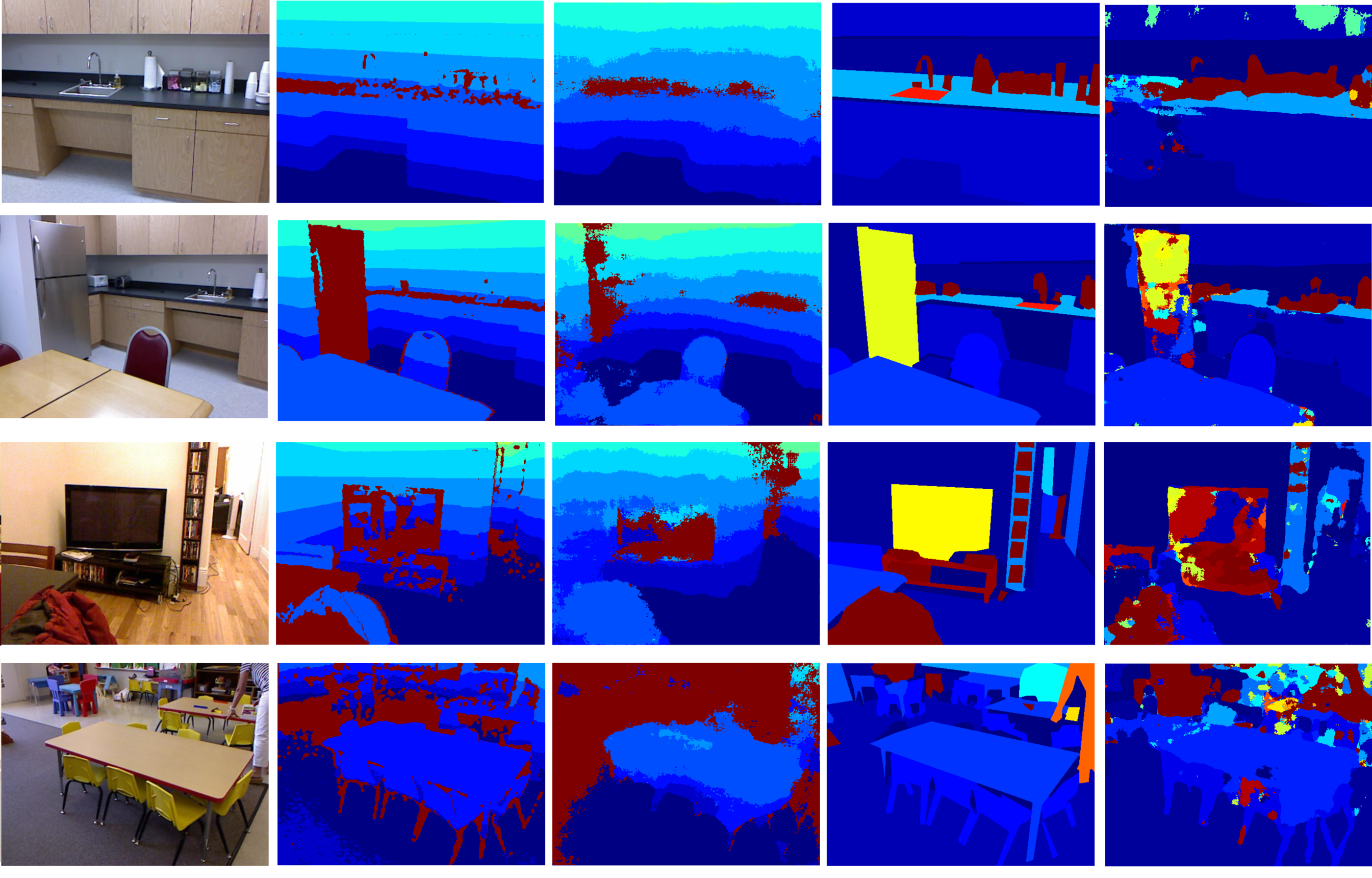}
\end{center}
   \vspace*{-3mm} 
   \caption{Qualitative evaluation of our proposed HN-network as well as the output of the semantic segmentation network. From left to right: Input RGB image, groundtruth height label, output height label, semantic groundtruth label, and output semantic label.}
\label{fig:qualitative}
\end{figure*}

\vspace{3pt}\noindent \textbf{Cross-Stitching: }
 We also conduct an experiment to fuse the HN-network with the semantic segmentation network. Instead of manually selecting the layers to share, our cross-stitch network learns to weigh the amount of sharing required between the two streams. The results are shown in Table \ref{tab:results_cross}, where we compare the final cross-stitched network to the non-stitched network and to the network pre-trained on ImageNet. We observe in our experiments that the final semantic segmentation accuracy increases without the need to manually select the number of layers to share and that cross-stitching is an essential component of its final performance. This represents a better way of transfer learning than basic initialization.

\vspace{3pt}\noindent \textbf{Qualitative Results: } Figure \ref{fig:qualitative} presents a set of  RGB images from NYUv2 with their corresponding groundtruth height labels, predicted height labels from HN-network, along with the semantic groundtruth labels and the predicted  semantic labels after cross-stitching. These images show that the current technique is capable of aptly generating semantic labels for RGB images along with height labels as a by-product. Also, the relation between the two tasks can be observed in areas like the ground, where a single height label corresponds to a single semantic label.  


\section{Conclusion}
We present a novel method to pre-train any semantic segmentation network without the need for expensive manual labeling. Our pre-training learns from object heights, which are automatically extracted from depth sensors. Extensive experiments show that our proposed pre-training is better than initializing with ImageNet, while using much less data and without requiring any manual labels. We also propose to fuse the pre-training with the semantic segmentation network to better differentiate between task specific and shared representations, leading to higher overall accuracy. Since our pre-training does not require manual annotation, it is easily scalable. Therefore, we suggest exploring such use of information to better improve the pre-training process to further improve performance over ImageNet pre-training.

{\small
\bibliographystyle{ieee_fullname}
\bibliography{egbib}
}

\end{document}